\documentclass[conference]{IEEEtran}

\usepackage{multicol}
\usepackage{diagbox}
\usepackage{multirow}
\usepackage{times}
\usepackage{amsmath,amssymb,amsfonts}
\usepackage{epsfig}
\usepackage{graphicx}
\usepackage{algorithmic}
\usepackage[lined,boxed,commentsnumbered]{algorithm2e}
\usepackage{bm}
\usepackage[bookmarks=false]{hyperref}
\usepackage{caption}
\usepackage{cite}
\usepackage{color}
\usepackage[export]{adjustbox}
\usepackage{graphicx}
\usepackage{subcaption}
\usepackage{url}
\usepackage{multirow}
\usepackage{gensymb}

\graphicspath{ {./figures_small/}} 
\newcommand{\rpm}{\raisebox{.2ex}{$\scriptstyle\pm$}}

\def\BibTeX{{\rm B\kern-.05em{\sc i\kern-.025em b}\kern-.08em
    T\kern-.1667em\lower.7ex\hbox{E}\kern-.125emX}}
\begin{document}

\title{Feature-less Stitching of Cylindrical Tunnel \\
}


\author{Ramanpreet~Singh~Pahwa$^{1}$,~Wei~Kiat~Leong$^{2}$,~Shaohui~Foong$^{3}$,~Karianto~Leman$^{1}$,~Minh~N.~Do$^{4}$
\thanks{$^{1}$R.~S.~Pahwa and K.~Leman are with Institute for Infocomm Research (I$^2$R), Singapore (e-mail: $\{$ramanpreet$\_$pahwa,~karianto$\}$@i2r.a-star.edu.sg). }
\thanks{$^{2}$W.~K.~Leong is with University of Glasgow, UK (e-mail: weikiat.leong87@gmail.com). }%
\thanks{$^{3}$S.~Foong is with University of Technology and Design (SUTD), Singapore (e-mail: foongshaohui@sutd.edu.sg). }%
\thanks{$^{4}$Minh~N.~Do is with the Department
of ECE, University of Illinois at Urbana-Champaign, IL, USA (e-mail: minhdo@uiuc.edu).}
}

\maketitle

\begin{abstract}
Traditional image stitching algorithms use transforms such as homography to combine different views of a scene. They usually work well when the scene is planar or when the camera is only rotated, keeping its position static. This severely limits their use in real world scenarios where an unmanned aerial vehicle (UAV) potentially hovers around and flies in an enclosed area while rotating to capture a video sequence. We utilize known scene geometry along with recorded camera trajectory to create cylindrical images captured in a given environment such as a tunnel where the camera rotates around its center. The captured images of the inner surface of the given scene are combined to create a composite panoramic image that is textured onto a 3D geometrical object in Unity graphical engine to create an immersive environment for end users.
\end{abstract}


\begin{section}{Introduction}
Aging infrastructure is becoming an increasing concern in the developed countries. There is a growing need for automatic or user-assisted assessment, diagnosis and fault detection of old structures such as sewage tunnels, bridges and roof-tops \cite{samiappan2016using, metni2007uav}. Some of these structures may also be inaccessible or too dangerous for human inspection. For example, manual inspection of deep tunnel networks is an extremely challenging and risky task due to the inaccessibility  and potentially hazardous environment contained in these tunnels. Due to the health risks involved, UAVs coupled with scene understanding techniques \cite{raman_tcsvt_3D_prop, raman_apsipa_3D_prop} provide a perfect choice  as they are compact and can be automated or controlled by a user to remotely capture the necessary information.

This paper builds towards imaging and inspection of the Deep Tunnel Sewerage System (DTSS). DTSS is a massive integrated project currently being developed by the Public Utilities Board (PUB) in Singapore to meet the country's long-term clean water needs through the collection, treatment, reclamation and disposal of used water from industries, homes and businesses \cite{dtss_cite}. These DTSS tunnels are covered with a corrosion protection lining (CPL) for protection. This paper aims towards automatically stitching the images collected by the UAV into a cylindrical panoramic view of the tunnel and render the tunnel in $3$D to inspect the physical conditions of the CPL as well as the structural integrity of the tunnel as a whole.

While UAVs provide a viable alternative for remote assessment of deep tunnels as they are unaffected by debris and sewage flow, they are primarily designed for high altitude aerial imagery and are not appropriate for short range detailed imaging of tunnel surfaces. An alternative is to attach a $360\degree$ camera in front of a UAV and capture the panoramic view of the tunnel. However, these $360\degree$ images have low resolution that are not suitable for fault detection. Moreover, most of these cameras are too heavy and/or consist of odd shapes that render them difficult to attach to a UAV. Instead, we can use a lightweight and high resolution camera. After performing calibration \cite{raman_icip2014, raman_MS, raman_PhD}, the camera can be mounted on a UAV. The camera rotates around the shaft of the UAV while it moves forward in a tunnel, in turn providing us with spiral-like images. 
\begin{figure}[t!]
\centering
\includegraphics[width=0.45\textwidth]{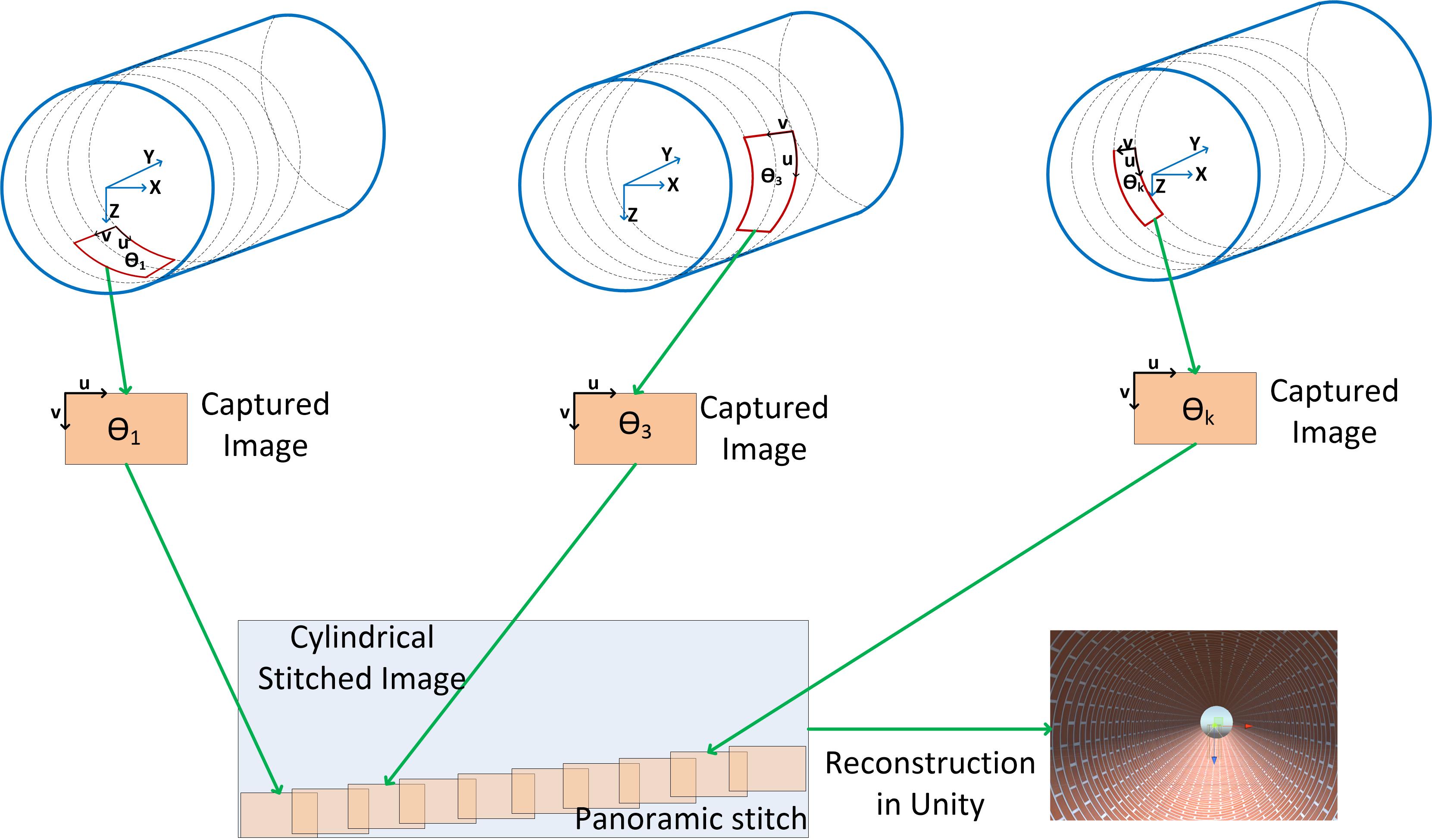}
\caption{ The UAV is assumed to be moving horizontally during the capturing process. This constant horizontal motion of the UAV coupled with the rotating camera results in a panoramic spiral image.}
\label{fig::rotating_moving_cam_stitch} 
\end{figure}

This paper presents a framework where we simulate a UAV to fly through a cylindrical sewage tunnel as shown in Fig.~\ref{fig::rotating_moving_cam_stitch}. We record the trajectory of the UAV moving in the scene and and integrate it with $2$D color images captured by a rotating camera to stitch the images using cylindrical projection. Thereafter, the stitched cylindrical images are textured on a tunnel-like $3$D object and displayed in Unity \cite{Unity2017} to assist users in visualization, remote inspection, and  fault detection of these tunnels.

In particular, we make the following contributions:
\begin{itemize}
\item A novel $360\degree$ revolving camera system is simulated in a virtual environment which, coupled with the maneuverability of the UAV, allows capturing high definition images of the tunnel surface efficiently in a spiraling motion. 
\item We develop a mathematical framework to combine recorded camera trajectory with known scene geometry of a tunnel to automatically combine the images captured and create a ``panoramic'' view of the tunnel.
\item A geometrical visualization framework is developed using Unity to assist the users in visualizing the tunnel and room-like geometry for inspection and fault detection.
\end{itemize}
\end{section}

\begin{section}{Related Work}\label{sec::related_work}
\noindent \textbf{Image Stitching}: A lot of work has been done in the computer vision \cite{JBlu_icassp11, ttng_ijcv12, ttng_iccv09, chu10compressive} and photogrammetry community \cite{szeliski2006image, zomet2006seamless, levin2004seamless, brown2007automatic} to perform image stitching. Traditional image stitching techniques rely on an underlying transform, usually a $3\times3$ affine or homographic matrix that maps  pixels from one coordinate frame to another. Typical image stitching techniques such as AutoStitch \cite{brown2008autostitch} assume the camera's location to be static, i.e. a pure camera rotation between captured images, or the captured scene to be roughly planar. In our panoramic spiral imaging system, the camera both rotates and translates while capturing the scene. This translation is often not negligible as compared to the distance of the tunnel surface to the camera. Moreover, the planar assumption of the scene is invalid for us since we capture  images of a cylindrical tunnel. Dornaika and Chung \cite{dornaika2004mosaicking} proposed a heuristic approach of piecewise planar patches to overcome this issue. Other recent methods \cite{zaragoza2013projective,zhang2014parallax,lin2011smoothly} propose a different strategy of aligning the images partially in order to find a good seam to stitch different images in the presence of parallax. However, these methods rely heavily on reliable feature detection and matching, which might be difficult in the tunnel environment. Furthermore, these methods do not exploit known geometry of the scene. 
\newline
\noindent \textbf{Structure-from-Motion: } Structure-from-Motion (SfM) refers to the recovery of $3$D structure of the scene from given images. A widely known application of SfM is where ancient Rome is reconstructed using Internet images \cite{agarwal2011building, frahm2010building}. SfM has made tremendous progress in the recent years \cite{kushal2012photo,  wu2013towards}. Since, we use a simulated environment for this work, we use the ground-truth camera pose recorded while capturing the dataset.
\newline
\noindent $\bf{360\degree}$ \textbf{cameras: } MIT Technology review \cite{Mit_2017_360cam} identified $360\degree$ cameras as one of the top ten breakthrough technologies of $2017$. Usually two or more spherical cameras are used to stitch images and videos in real-time \cite{VIRB_360,insta_360} or offline \cite{mi_360,rylo_360}. However, the current technology suffers from extensive motion blur and low resolution. Expensive and dedicated hardware is required to capture and post-process high resolution video \cite{Kasahara_2015_jackin}. To our knowledge, the best resolution for $360\degree$ video capture is provided by VIRB $360$ \cite{VIRB_360} at $3,840$ pixels per $360\degree$ resolution at $30$ fps. This is not sufficient for anomaly inspection and identifying faults in sewage tunnel linings. We intend to use a light-weight GoPro camera \cite{gopro4_url} for our future data collection which will provide upto $7,500$ pixels per $360\degree$ resolution at $60$ fps. This will result in twice the resolution and twice the frame rate compared to the best solution available currently in the market. 

\end{section}
\begin{section}{Cylindrical Projection}\label{sec::cyl_projection} 
\begin{figure}[t!]
\centering
\includegraphics[width=0.45\textwidth]{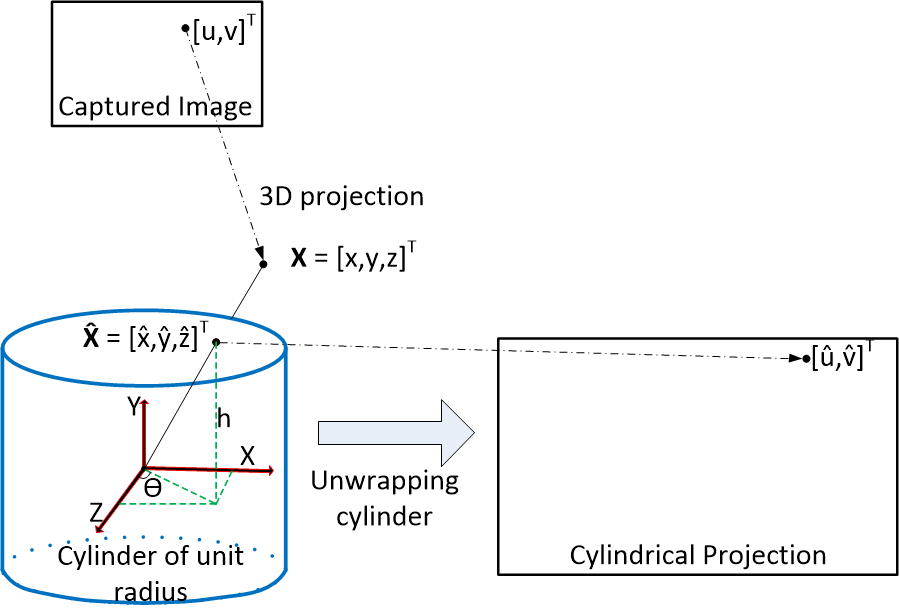}
\caption{Cylindrical projection - A $3$D point is projected onto a cylinder with unit radius to obtain cylindrical coordinates. These cylindrical coordinates are then eventually flattened out onto a planar image.}
\label{fig::cyl_projection} 
\end{figure}

In this section, we set up the cylindrical projection that models spiral imaging of tunnel surfaces. As shown in Fig.~\ref{fig::cyl_projection}, a generic $2$D pixel of an acquired image, $[u,v]^\intercal$, can be projected to a $3$D point $\bm{x} = [x,y,z]^\intercal$ using a camera's intrinsic projection parameters - focal length, $f$, and optical center ($[c_x, c_y]^\intercal$) - as follows:
\begin{align}
\bm{x} = \begin{bmatrix} x  \\ y \\ z \end{bmatrix} = \lambda \bm{K}^{-1} \begin{bmatrix} u  \\ v \\ 1 \end{bmatrix}  = \lambda
\begin{bmatrix} \frac{u-c_x}{f}  \\ \frac{v-c_y} {f}\\ 1 \end{bmatrix}  ,
\end{align}
where $\bm{K}$ represents the internal calibration matrix of the camera and $\lambda$ refers to the pixel's depth. This $3$D point is projected onto a unit cylinder as follows:
\begin{align}  \label{eq::3D_cyl_coords}
\begin{bmatrix} \hat{x}  \\ \hat{y} \\ \hat{z} \end{bmatrix} = \frac{1}{\sqrt{ x^2+z^2}}\begin{bmatrix} x  \\ y \\ z \end{bmatrix} = \begin{bmatrix} \sin\theta  \\ h \\ \cos\theta \end{bmatrix}  ,
\end{align}
where, angle, $\theta$, and height, $h$, are two parameters required to represent a $3$D point lying on a unit cylinder. The unit cylinder can be unwrapped onto a planar image as shown in Fig.~\ref{fig::cyl_projection} as follows:
\begin{align}\label{eq::x_ftheta}
\begin{bmatrix} \hat{u} \\ \hat{v} \end{bmatrix} = \begin{bmatrix} f \dot{•} \theta + c_x  \\ f\dot{•} h + c_y \end{bmatrix}  .
\end{align}

An example of cylindrical projection of a $2$D planar image captured by a camera is shown in Fig.~\ref{fig::cylindrical_projection}.
\begin{figure}[t!]
\centering
\begin{subfigure}[b]{0.225\textwidth}
		\includegraphics[width=\textwidth]{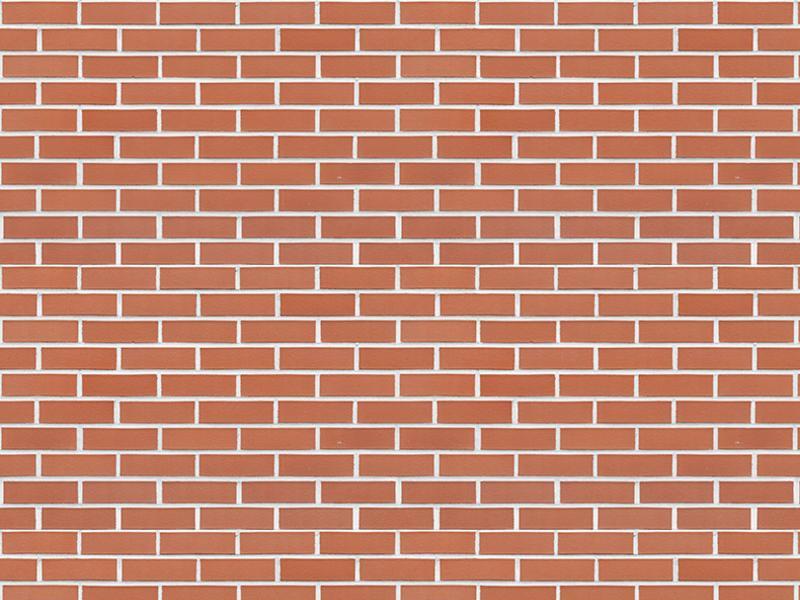}
		\caption{}
\end{subfigure}
    \begin{subfigure}[b]{0.225\textwidth}
		\includegraphics[width=\textwidth]{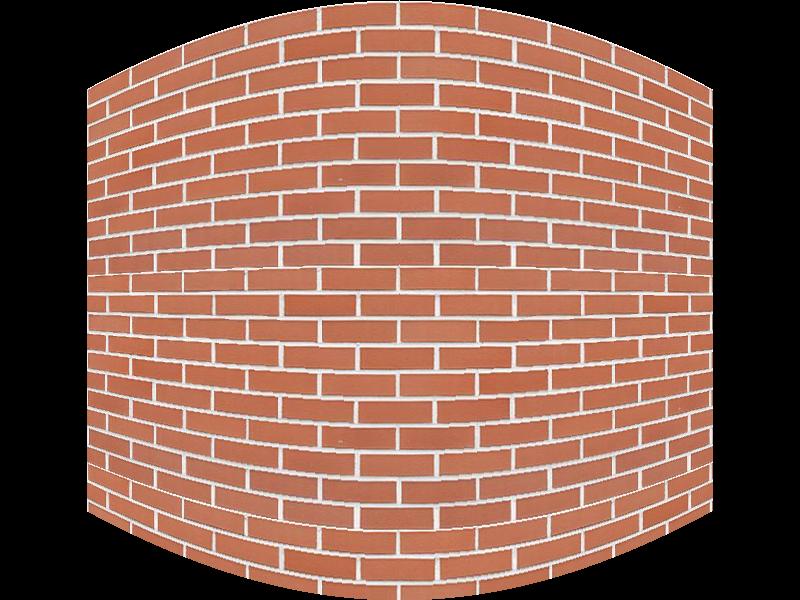}
		\caption{}
	\end{subfigure} 
\caption{ (a) A sample brick image. (b) Image obtained after cylindrical projection.}
\label{fig::cylindrical_projection} 
\end{figure}
\end{section}

\begin{section}{Image Stitching} \label{sec::our_algorithm}
We use Unity to generate our simulated dataset. A cylindrical hollow cylinder, resembling a tunnel, is created in Blender \cite{blender_cite} and a texture is applied to it in Unity. The camera is placed near the center of tunnel and is rotated a certain degree every few milliseconds. The benefits of taking an image after the camera rotates a certain pre-defined rotation is to avoid issues such as motion blur and rolling shutter noise that would occur if we capture a video sequence in real environment.

As we use a simulated environment, we can generate the ground-truth measurements for the movement of the camera in the tunnel. We consider these rotation and translation  measurements of the camera as the final camera pose for our framework. The image capturing and stitching design is shown in Fig.~\ref{fig::unity_sim_sample}.

Using the camera pose enables us to project and transform the $2$D pixel information per image into world coordinate frame. Let $\bm{X}_w$ represent the $3$D points, per frame, in world coordinate frame. A global $i^{th}$ point $\bm{x}_w = \bm{X}_w^i$ lying on the cylindrical tube of radius $r$ can be represented by two parameters - $\theta^i$ and $h^i$ as follows:
\begin{align}
\bm{X}_w^i    =  \bm{x}_w = \begin{bmatrix} r \sin\theta^i  \\ h^i \\ r\cos\theta^i \end{bmatrix}
\end{align}
Let $\bm{X}_c$ represent the $3D$ points in camera frame of reference for every image captured by the rotating camera. Every $i^{th}$ pixel (=$[u^i,v^i]^\intercal$) can be projected onto $3$D as:
\begin{align}
\bm{X}_c^i    =  \bm{x}_c =\frac{z_c^i}{f} \begin{bmatrix} u^i-c_x   \\ v^i-c_y \\ f \end{bmatrix}
\end{align}
where $z_c^i$ refers to the depth of $i^{th}$ pixel which is unknown. $\bm{X}_c$ and $\bm{X}_w$ are related by:
\begin{align}\label{eq::Xw}
\bm{X}_w    =  \bm{R}\bm{X}_c + \bm{t}
\end{align}
where $\bm{R}$ is a $3\times3$ orthonormal rotation matrix and $\bm{t} = [t_x,t_y,t_z]^\intercal$ represents the translation of the UAV in world coordinate frame. Initially the UAV is assumed to be at the center of the tunnel ($\bm{t} = \vec{0}$). As the UAV moves horizontally across the tunnel, $t_y$ increases. $t_x$ and $t_z$ denote the deviation of the UAV from center of the tunnel in $\vec{x}$ and $\vec{z}$ directions respectively.
\begin{figure}[t!]
\centering
	\begin{subfigure}[b]{0.225\textwidth}
		\includegraphics[width=\textwidth]{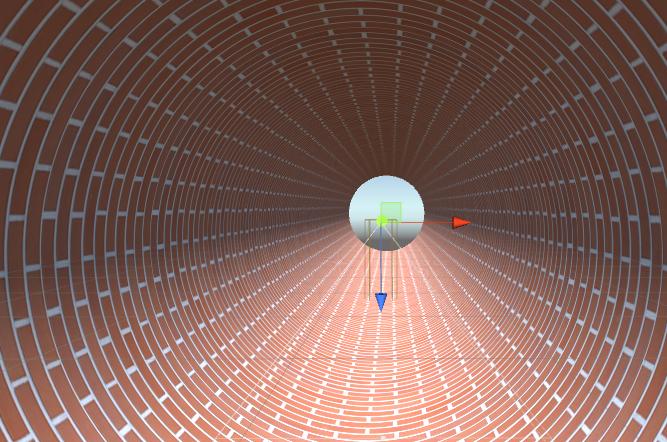}
		\caption{$\theta = 0 \degree$}
	\end{subfigure}  
	\begin{subfigure}[b]{0.225\textwidth}
		\includegraphics[width=\textwidth]{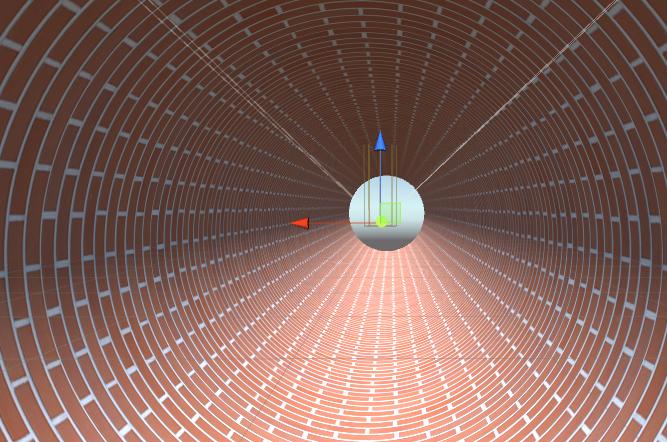}
		\caption{$\theta = 180 \degree$}
	\end{subfigure}  
	\begin{subfigure}[b]{0.225\textwidth}
		\includegraphics[width=\textwidth]{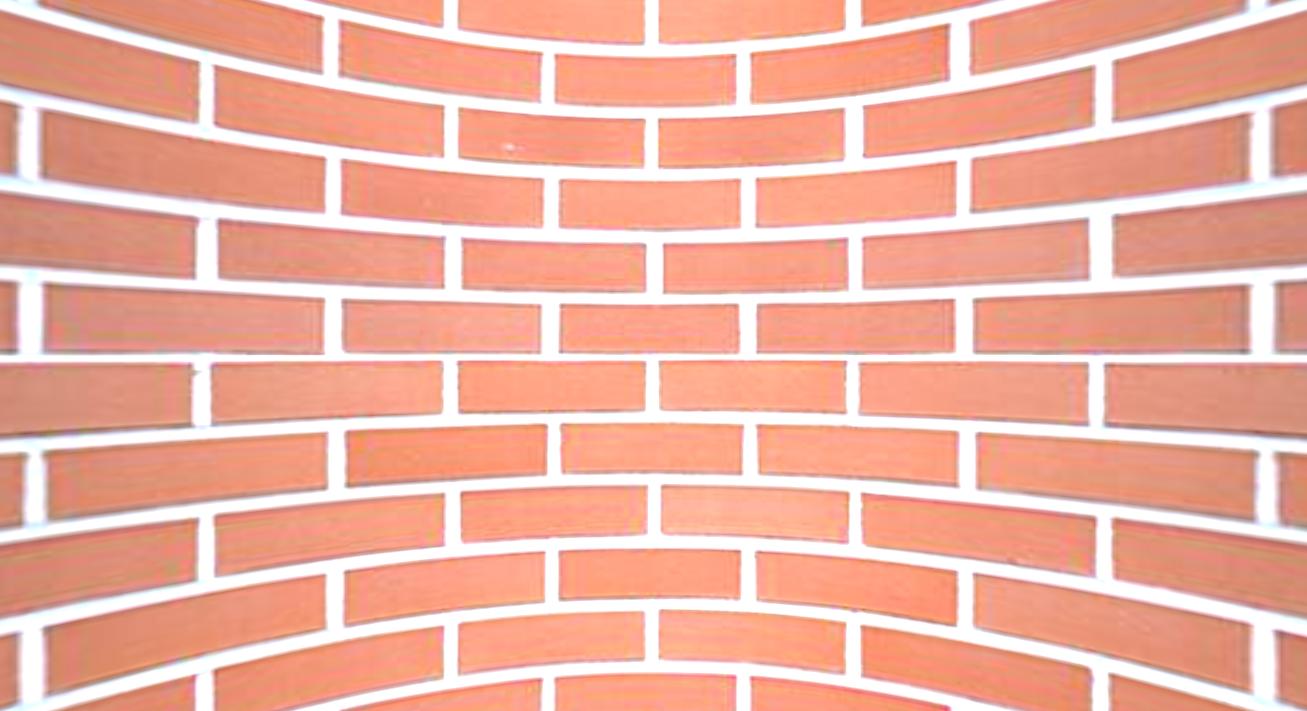}
		\caption{Image captured at $\theta = 0 \degree$}
	\end{subfigure}  
	\begin{subfigure}[b]{0.225\textwidth}
		\includegraphics[width=\textwidth]{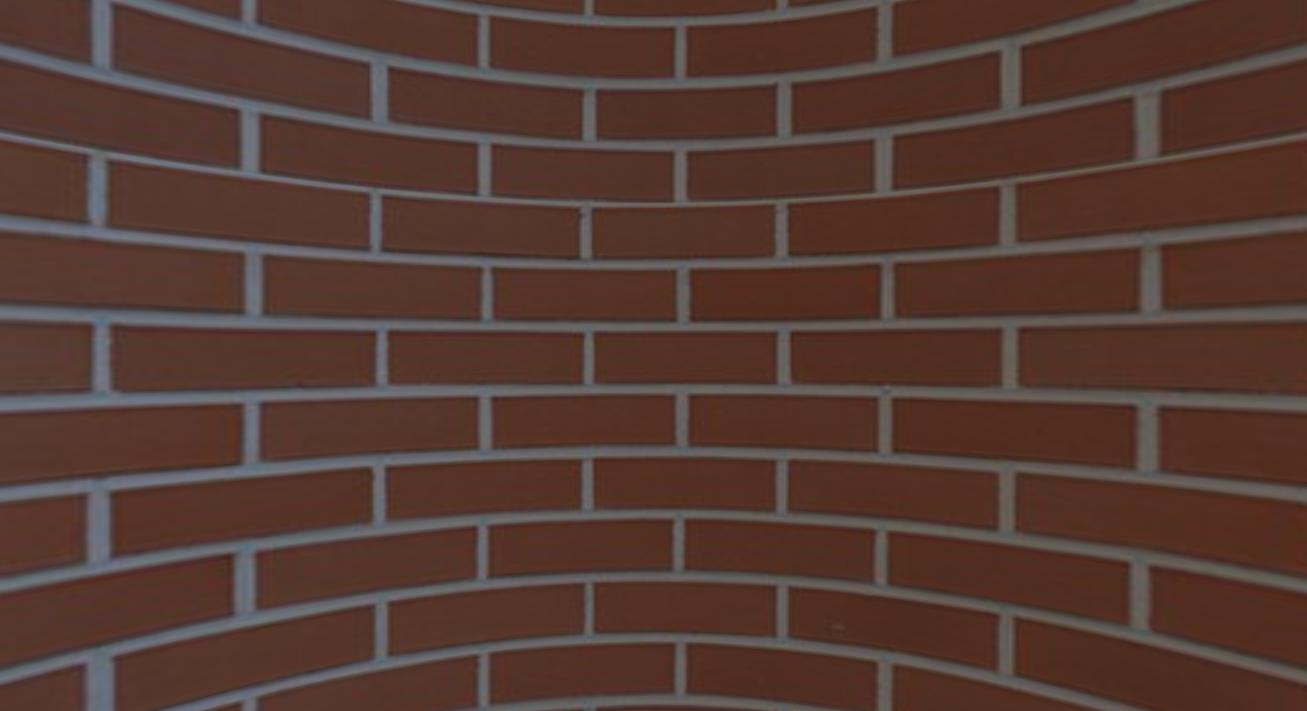}
		\caption{Image captured at $\theta = 180 \degree$}
	\end{subfigure}  
  \caption{Top row shows the unity simulation when camera is rotated by $0 \degree$ and $180 \degree$ with respect to y axis (green arrow). Bottom row displays the images captured by the virtual camera at these two angles respectively.}
\label{fig::unity_sim_sample}
\end{figure}
\begin{align*}
\bm{R}    =   \begin{bmatrix} r_{11} & r_{12} & r_{13} \\ r_{21} & r_{22} & r_{23}  \\ r_{31} & r_{32} & r_{33}  \end{bmatrix} \qquad \bm{t} =\begin{bmatrix} t_{x} \\ t_{y}   \\ t_{z} \end{bmatrix} 
\end{align*}

For every $i^{th}$ pixel, we obtain three equations for three parameters - $\{z_c^i$, $\theta^i$, $h^i\}$. The two equations involving $\theta$ can be deterministically solved as follows:
\begin{align}\label{eq::rsin}
r \sin\theta^i  &=  (\bm{R}\bm{x}_c)(1) +t_x \\
  &= \frac{z_c^i}{f}(r_{11}(u^i-c_x) + r_{12}(v^i-c_y) + r_{13}f) + t_x \nonumber
\end{align}
\begin{align}\label{eq::rcos}
r \cos\theta^i  &=  (\bm{R}\bm{x}_c)(3) +t_z \\
&= \frac{z_c^i}{f}(r_{31}(u^i-c_x) + r_{32}(v^i-c_y) + r_{33}f) + t_z  \nonumber
\end{align}
where $(\bm{R}\bm{x}_c)(1)$ and $(\bm{R}\bm{x}_c)(3)$ represent the first and third element of the column vector $\bm{R}\bm{x}_c$. Let
\begin{align}\label{eq::c123}
\begin{bmatrix} c_1 \\c_2 \\c_3 \end{bmatrix} = \frac{1}{f}\begin{bmatrix} (r_{11}(u-c_x) + r_{12}(v-c_y) + r_{13}f)\\ (r_{21}(u-c_x) + r_{22}(v-c_y) + r_{23}f) \\ (r_{31}(u-c_x) + r_{32}(v-c_y) + r_{33}f) \end{bmatrix}
\end{align}
Solving for $z_c$ using Eqs.~(\ref{eq::Xw},\ref{eq::rsin},\ref{eq::rcos},\ref{eq::c123}), we obtain:
\begin{align}
r^2 = z_c^2(c_1^2+c_3^2)+2z_c(c_1 t_x + c_3 t_z) + t_x^2 + t_z^2 
\end{align}
Thus,
{\small
\begin{align} \label{eq::z_c}
z_c = \frac{-(c_1 t_x + c_3 t_z) \rpm  \sqrt{(c_1 t_x + c_3 t_z) - (c_1^2+c_3^2)(t_x^2 + t_z^2-r^2)} }{(c_1^2+c_3^2)}
\end{align}
}
Thereafter, we can compute $\theta$ by finding an intersection of the following two solutions for Eqs.~(\ref{eq::rsin},\ref{eq::rcos}):
{\small
\begin{align} \label{eq::r_sintheta}
\theta = & \{ \arcsin(\frac{c_1 z_c+t_x}{r}),\pi - \arcsin(\frac{c_1 z_c+t_x}{r}) \}  \quad \cap  \nonumber \\ & \{ \arccos(\frac{c_3 z_c+t_z}{r}),2\pi - \arccos(\frac{c_3 z_c+t_z}{r}) \}
\end{align}
}
Thereafter, we can also estimate the $y$ component of $\bm{x_c}$ by:
\begin{align}\label{eq::hi}
h =  c_2 z_c +t_y
\end{align}
Once we compute the world coordinate of every pixel's location,  we can obtain the cylindrical projection for it as described in Sec.~\ref{sec::cyl_projection}.
\end{section}

\begin{section}{Experimental Results}\label{sec::exper_results}
In this section, we perform synthetic experiments and compare our results in both noiseless and noisy scenarios. We also discuss our Unity framework and display a few examples of the rendered cylindrical scene in Unity for visualization. Unity is a game engine used mainly to create platform independent video game applications. However, it also provides us with a good set of tools to provide an immersive $3$D visualization for various purposes. The cylinder of radius, $r=3$m, is positioned such that the geometrical center of the cylinder is located at $[0,0,0]^\intercal$. We texture-mapped a brick wall  with a downward facing light source onto the inner face of the cylinder for visualization purposes. The light source is fixed to look vertically down for our experimental evaluation. Hence, images captured around $0 \degree$ rotation are brightly lit while the images captured around $180 \degree$ appear dark due to lack of illumination. A few images captured by the virtual camera are shown in Fig.~\ref{fig::unity_sim_sample}(c-d), ~\ref{fig::unity_sim_stationary_cam}(a-d). Let us denote the cylindrical image to be synthesized as $\bm{X}^{2D}_w$. 
\newline
\noindent \textbf{Stationary Camera Positioned at Center}: In our first experiment, we positioned the camera at the center of a cylindrical tunnel. The camera is held stationary throughout this experiment and only rotated rotated by $\beta = 30 \degree$ per frame and images are captured consequently. An example of this process is shown in Fig.~\ref{fig::unity_sim_stationary_cam}. It takes $12$ images to complete the full $360\degree$ rotation. Thereafter, the images are stitched together as described in Sec.~\ref{sec::our_algorithm}. In this experiment:
\resizebox{.9\linewidth}{!}{
  \begin{minipage}{\linewidth}
  \begin{align}
\bm{R}    =   \begin{bmatrix} \cos(\beta k) & 0 & \sin (\beta k) \\ 0 & 1 & 0  \\ -\sin(\beta k) & 0 & \cos(\beta k)  \end{bmatrix}; \bm{t} =\begin{bmatrix} 0 \\ 0  \\ 0 \end{bmatrix} \forall k = \{0, 1, \hdots , 11 \}
\end{align}
  \end{minipage}
}
\newline \newline \noindent For each image captured, we use Eqs.~\ref{eq::r_sintheta} and \ref{eq::hi} to project each pixel onto the cylindrical stitched image  $\bm{X}^{2D}_w$. However, performing this ``forward warping" may leave some holes in the stitched image. Thus, instead of performing forward warping, we use the four corners of each image to obtain the forward warped boundary. Thereafter, for every pixel inside this boundary of $\bm{X}^{2D}_w$, we perform ``inverse warping" to obtain its pixel location and intensity information in $\bm{X}^i_c$. The fully stitched image is shown in Fig.~\ref{fig::unity_sim_stationary_cam}(e). We observe that the images align perfectly with each other and all the ``curved" bricks are straightened after performing cylindrical projection.
\begin{figure}[t!]
\centering
	\begin{subfigure}[b]{0.11\textwidth}
		\includegraphics[width=\textwidth]{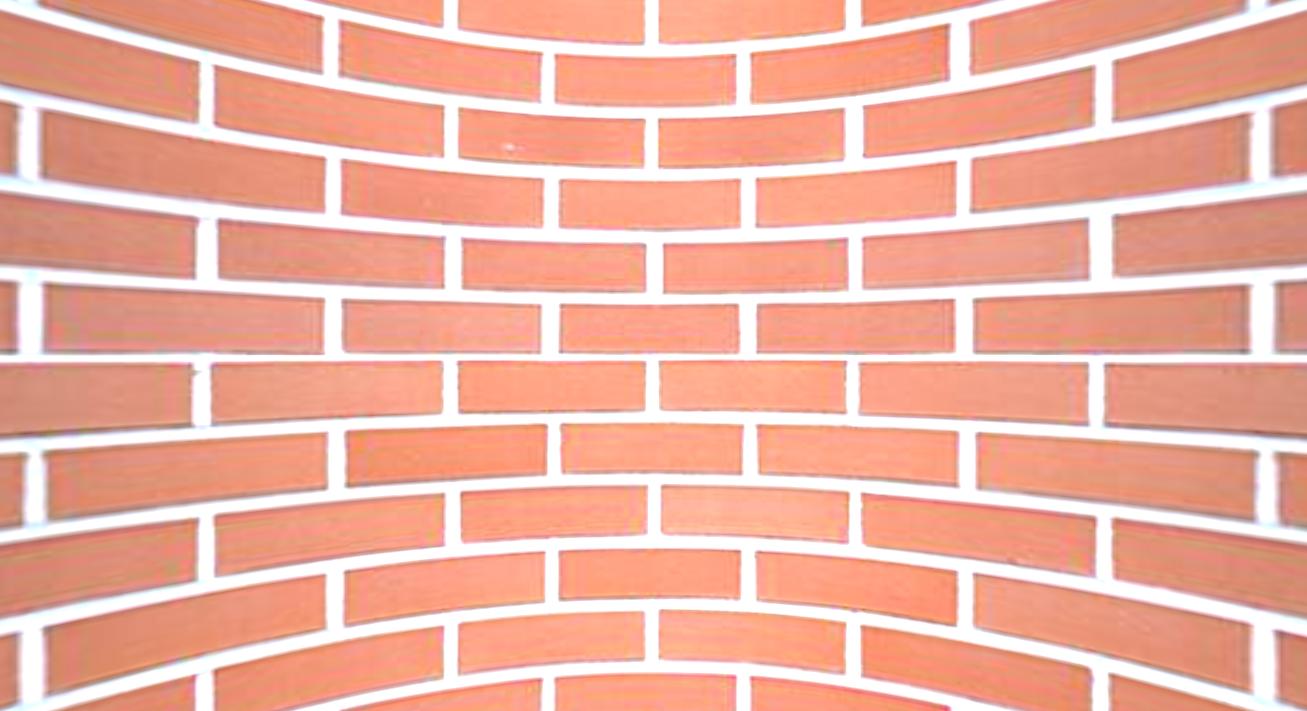}
		\caption{$\theta = 0 \degree$}
	\end{subfigure}  
	\begin{subfigure}[b]{0.11\textwidth}
		\includegraphics[width=\textwidth]{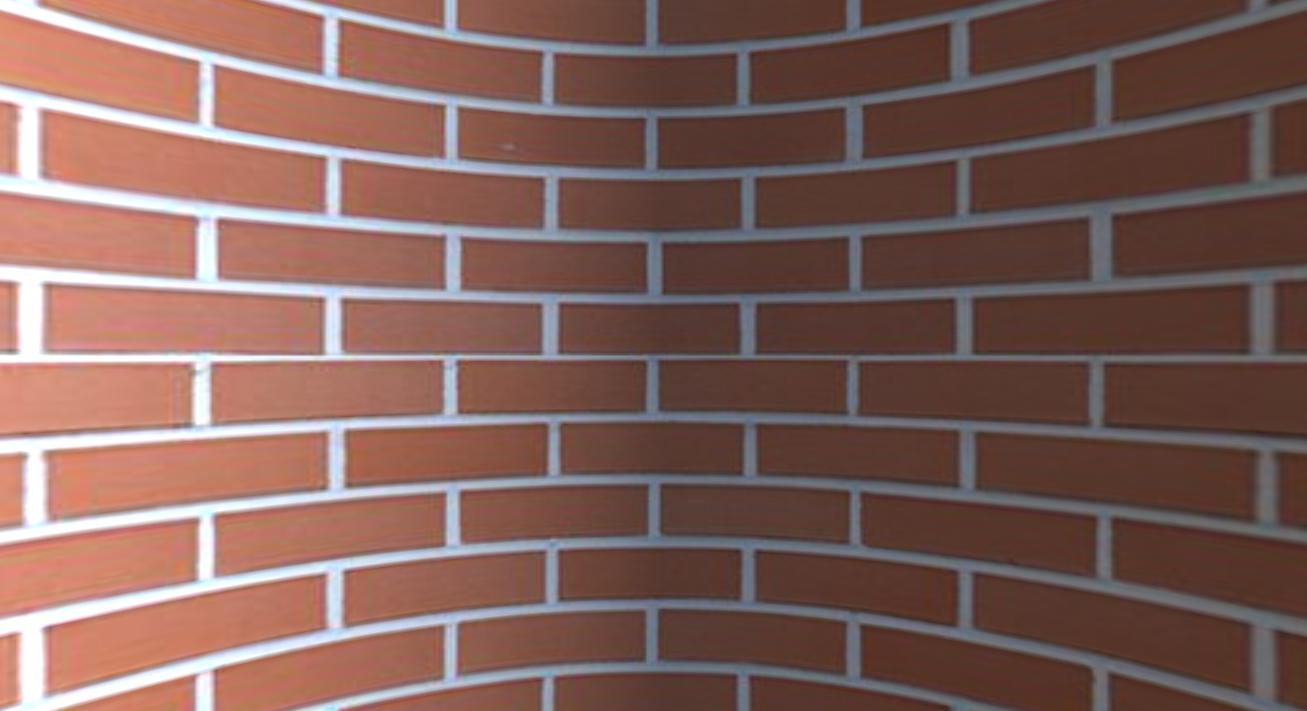}
		\caption{$\theta = 90 \degree$}
	\end{subfigure}  
	\begin{subfigure}[b]{0.11\textwidth}
		\includegraphics[width=\textwidth]{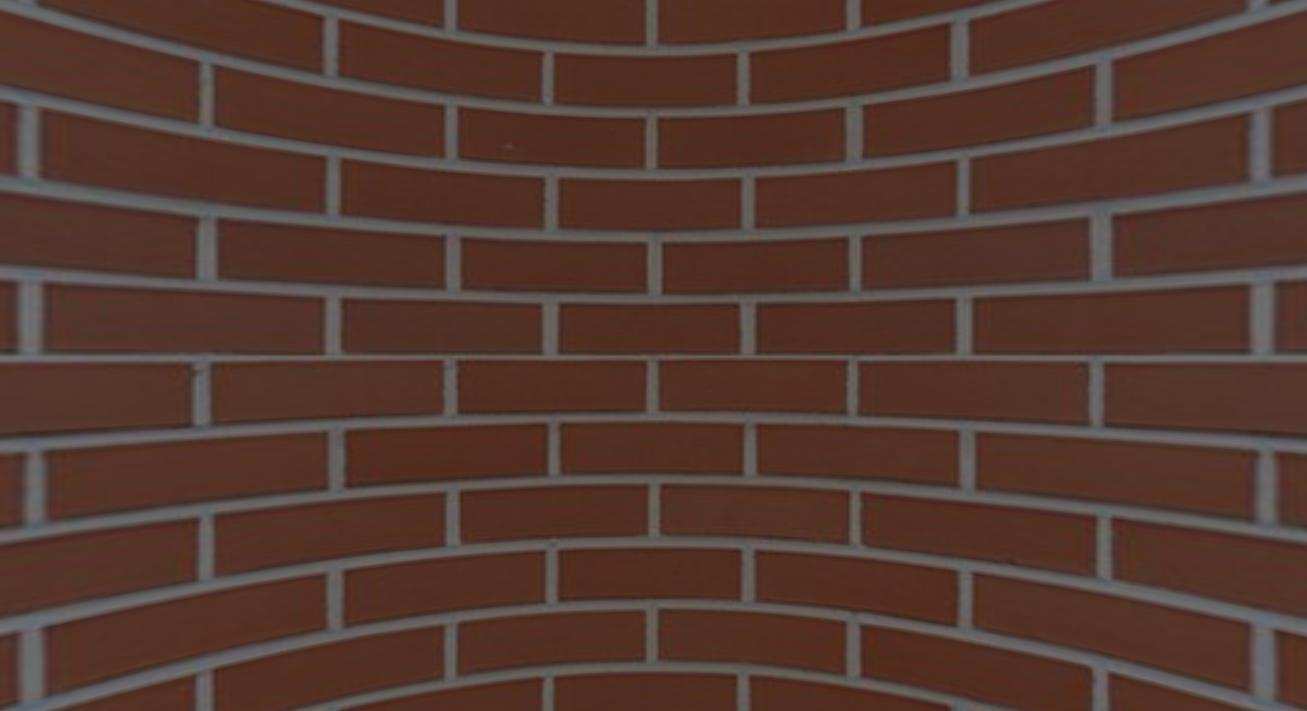}
		\caption{$\theta = 180 \degree$}
	\end{subfigure}  
	\begin{subfigure}[b]{0.11\textwidth}
		\includegraphics[width=\textwidth]{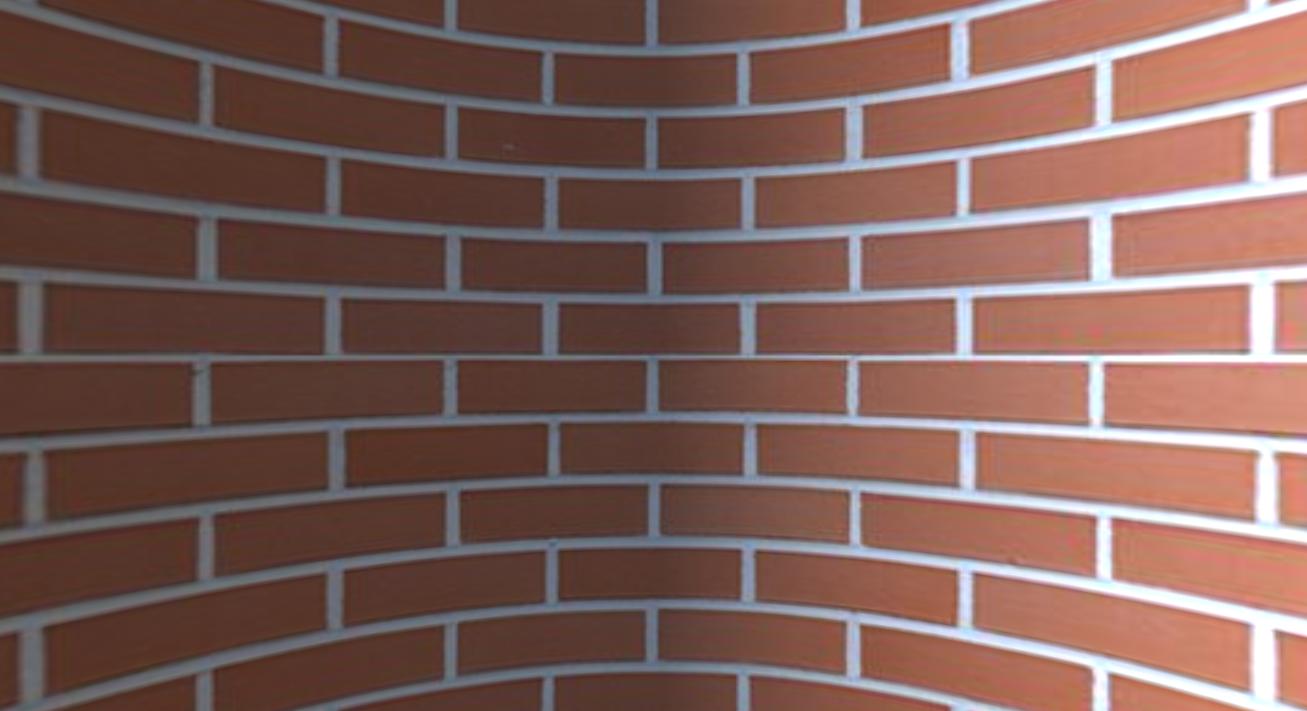}
		\caption{$\theta = 270 \degree$}
	\end{subfigure}  
		\begin{subfigure}[b]{0.45\textwidth}
		\includegraphics[width=\textwidth]{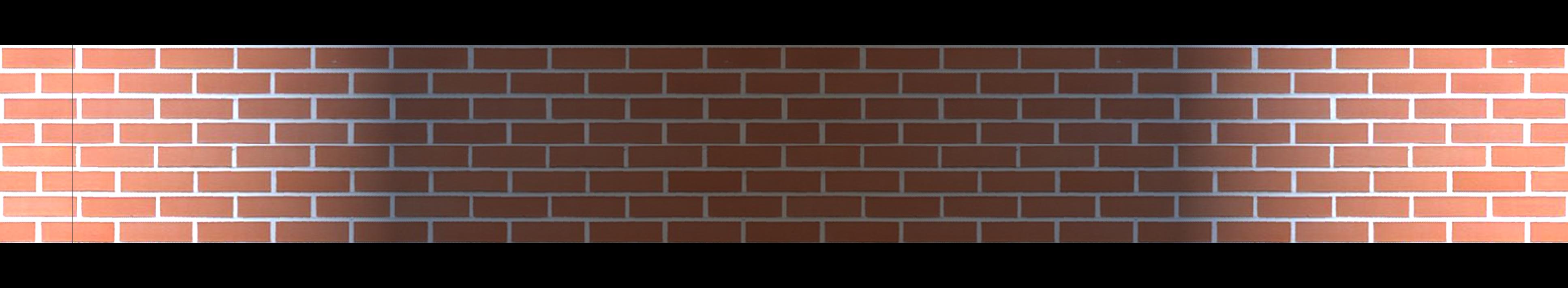}
		\caption{Stitched image}
	\end{subfigure}  
  \caption{Top row displays the images captured from unity when the camera is rotated from $0$ to $360$ degrees. Bottom row displays the stitched image obtained after our stitching process.}
\label{fig::unity_sim_stationary_cam}
\end{figure}
\newline
\noindent \textbf{Stationary Camera Positioned off Center}: Under ideal conditions, the camera should be positioned at the center of cylindrical tunnel. However, in reality this will not happen as UAVs tend to hover around and might have difficulty maneuvering to the center of the cylinder before each image is captured. We move the camera off center, i.e. $\bm{t} = [0.5m,0m,0.5m]^\intercal$.  The camera is held stationary throughout this experiment and only rotated by $\beta = 30 \degree$ per frame and images are captured consequently. 

A cylindrical image is a $360\degree$ view of the scene around the camera flattened on a planar image. Thus, even though the camera is off center, we can still view what the cylindrical projection of the scene looks like assuming the UAV's current position to be the center of the tunnel as shown in Fig.~\ref{fig::unity_sim_stationary_cam_off_center}(a). While the stitching is perfect, the straight lines (bricks) are no longer straight and we can see the zoom in and zoom out effect when the camera is far or near to the tunnel boundary respectively. We use Eqs.~\ref{eq::r_sintheta} and \ref{eq::hi} to synthesize what each image would look like if the camera was positioned at the center of the tunnel.  The fully stitched image is shown in Fig.~\ref{fig::unity_sim_stationary_cam_off_center}(b). We observe that the images align perfectly with each other and all the ``curved" bricks are straightened after performing cylindrical projection.
\begin{figure}[t!]
\centering
	\begin{subfigure}[b]{0.45\textwidth}
		\includegraphics[width=\textwidth]{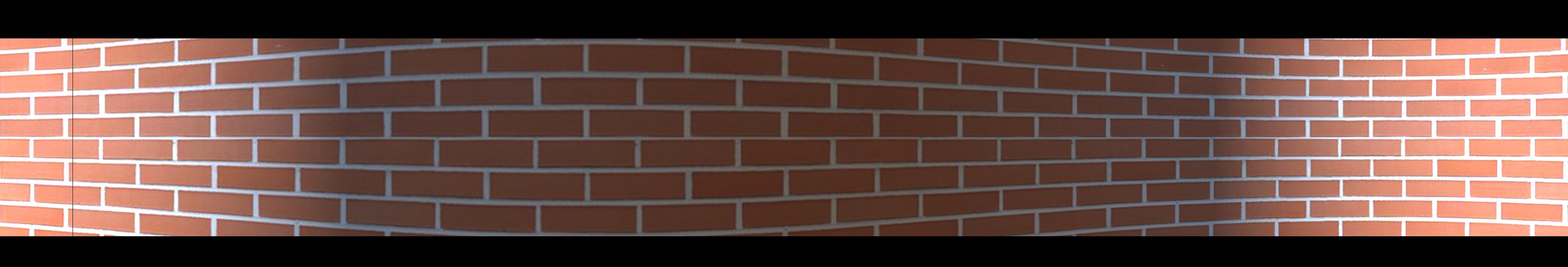}
		\caption{}
	\end{subfigure}  
	\begin{subfigure}[b]{0.45\textwidth}
		\includegraphics[width=\textwidth]{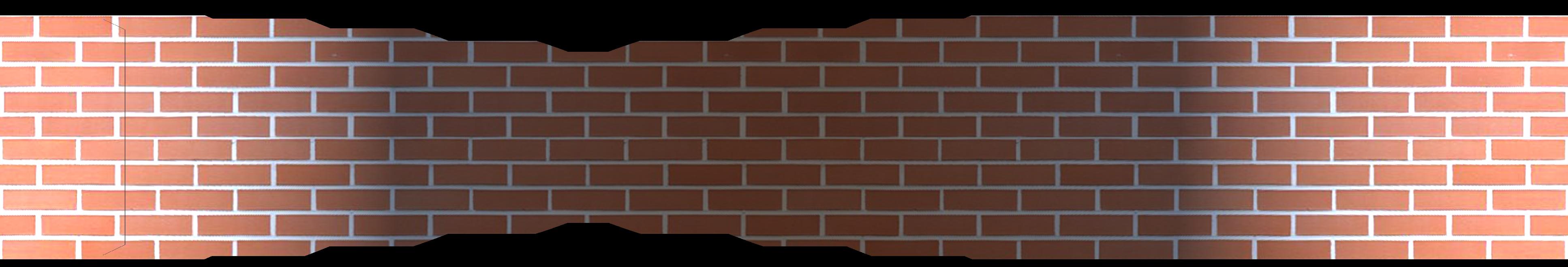}
  \caption{}
  	\end{subfigure}  
\caption{(a) Stitched cylindrical image without accounting for camera location off center. (b)  Stitched cylindrical image after accounting for camera's location off center.}
\label{fig::unity_sim_stationary_cam_off_center}
\end{figure}

\noindent \textbf{Freely moving camera in a simulated tunnel}: In our last experiment, we aim to simulate UAV movements in real-world conditions. A UAV is expected to suffer from jitters and sideways movements while it tries to balance itself and move forward in the tunnel. This means that the camera's movement and rotation per image given by IMU may be unreliable for our stitching purposes. 
\begin{figure}[t!]
\centering
\begin{subfigure}[b]{0.45\textwidth}
		\includegraphics[width=\textwidth]{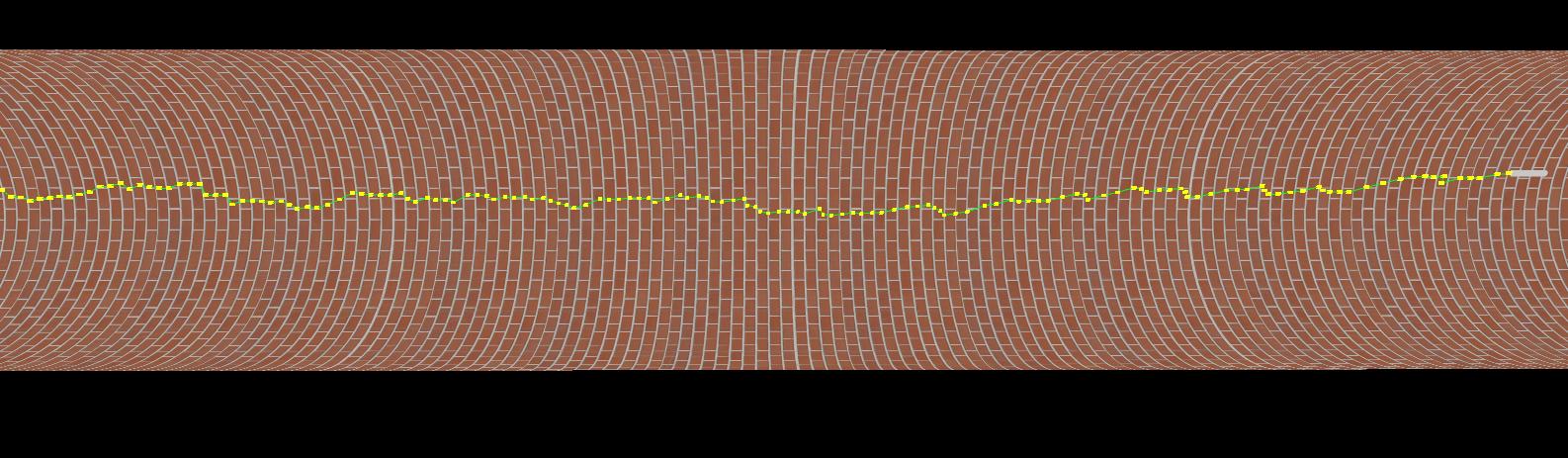}
  \caption{Simulated camera trajectory inside the tunnel.}
  	\end{subfigure}
			\begin{subfigure}[b]{0.45\textwidth}
		\includegraphics[width=\textwidth]{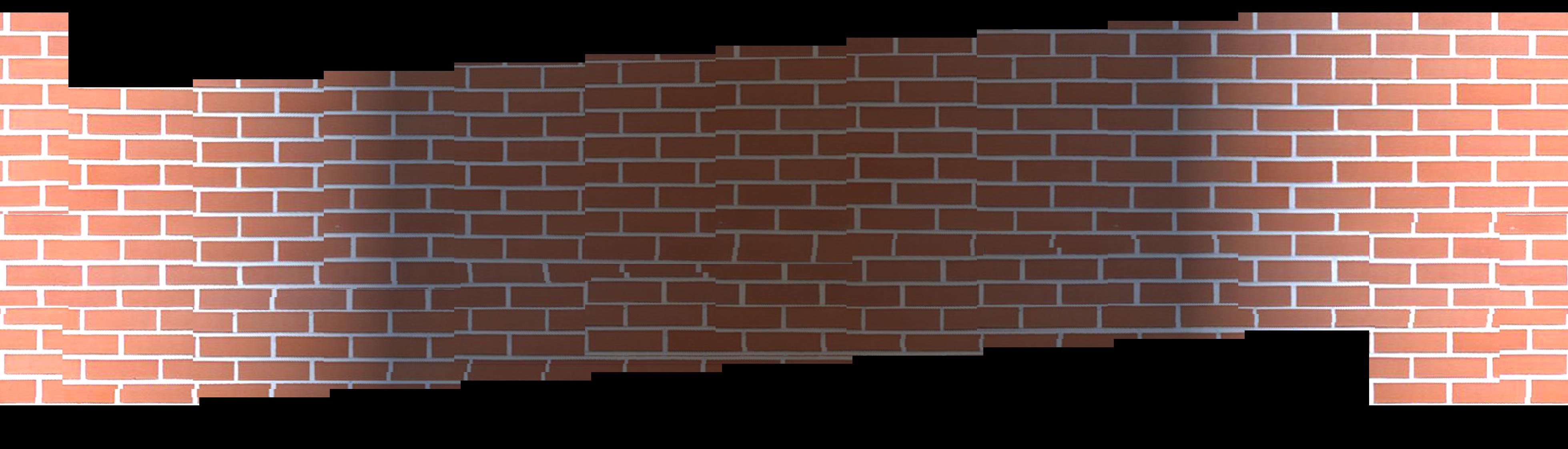}
  \caption{Using baseline pre-estimated camera pose.}
  	\end{subfigure}
  	\begin{subfigure}[b]{0.45\textwidth}
		\includegraphics[width=\textwidth]{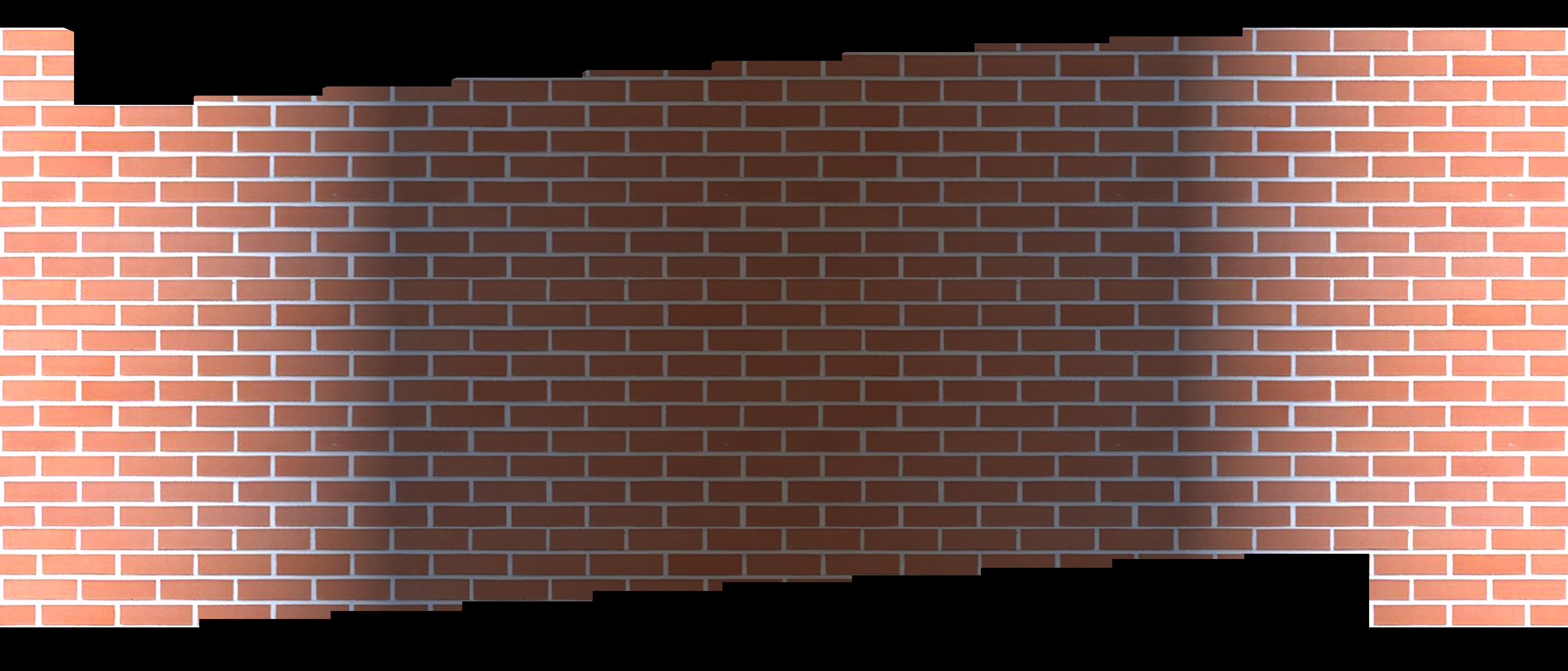}
  \caption{Using groundtruth camera pose, four rotations.}
  	\end{subfigure}
  	\begin{subfigure}[b]{0.45\textwidth}
		\includegraphics[width=\textwidth]{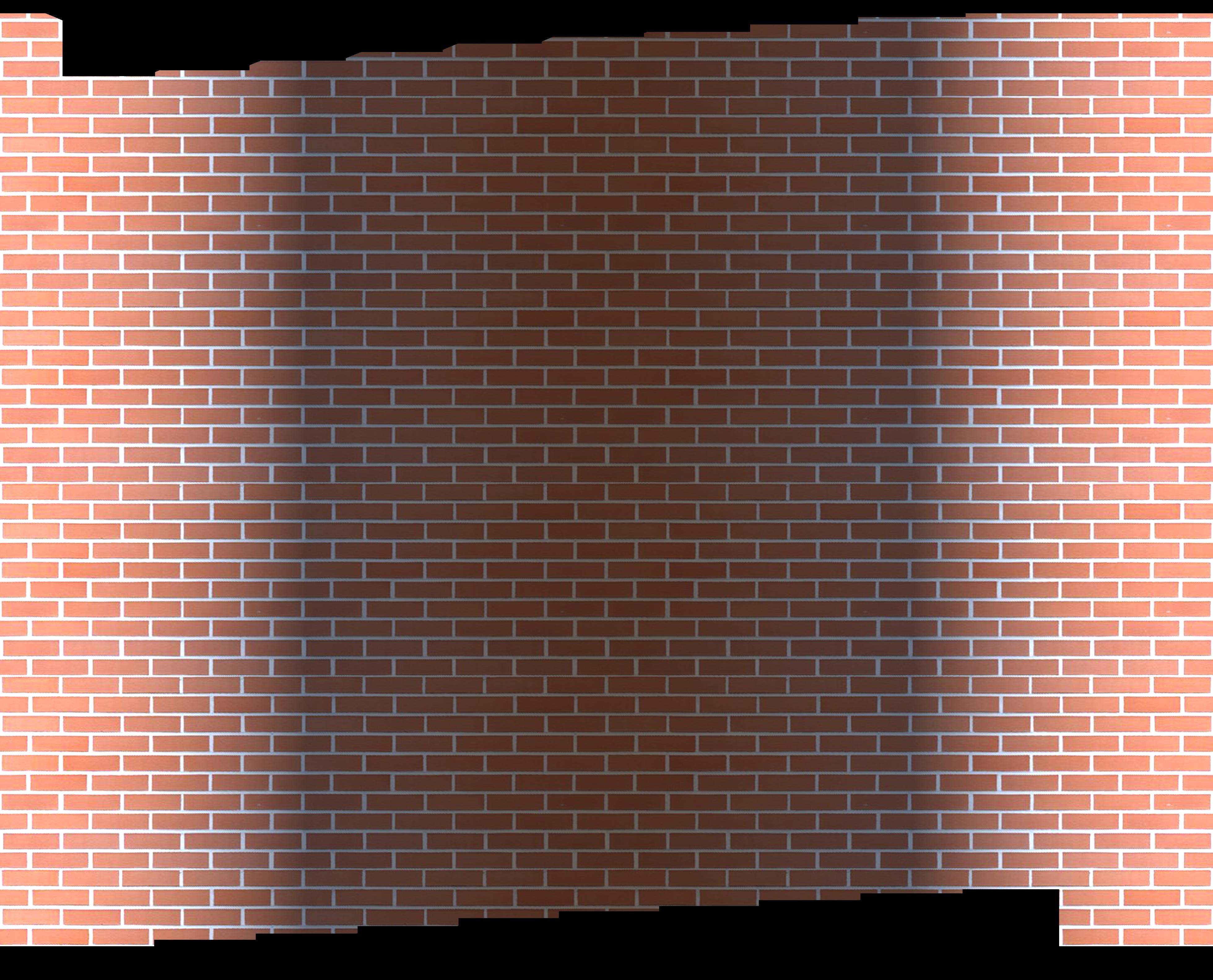}
  \caption{Using groundtruth camera pose, ten rotations.}
  	\end{subfigure}
  \caption{We add random jitter and movement to the camera after each image capture. Using baseline rotation and translation leads to an extremely inaccurate and incoherent stitch. 
\label{fig::SfM_stitch_seattle}}
\end{figure}
We simulate a tunnel of radius $3$ m. We initially positioned the camera at the center of the cylindrical tunnel. Our baseline movement and rotation in the three orthogonal directions are $\bm{t} = [0,0.10,0]^\intercal$ and $\bm{r} =[0,0.524,0]^\intercal$ respectively. This implies that we expect the ideal movements in the tunnel to be $10$ cm horizontally forward (y direction) with a $30\degree$ rotation across y axis. We add Gaussian noise with zero mean and standard deviation of $[2,2,3]^\intercal$ to our translation and Gaussian noise with zero mean and standard deviation of $2\degree$ to camera rotation per image. We run the simulation till the camera completes ten rotations and record the groundtruth translation and rotation of the camera per frame.  
\begin{figure}[t!]
\centering 
	\begin{subfigure}[b]{0.22\textwidth}
		\includegraphics[width=\textwidth]{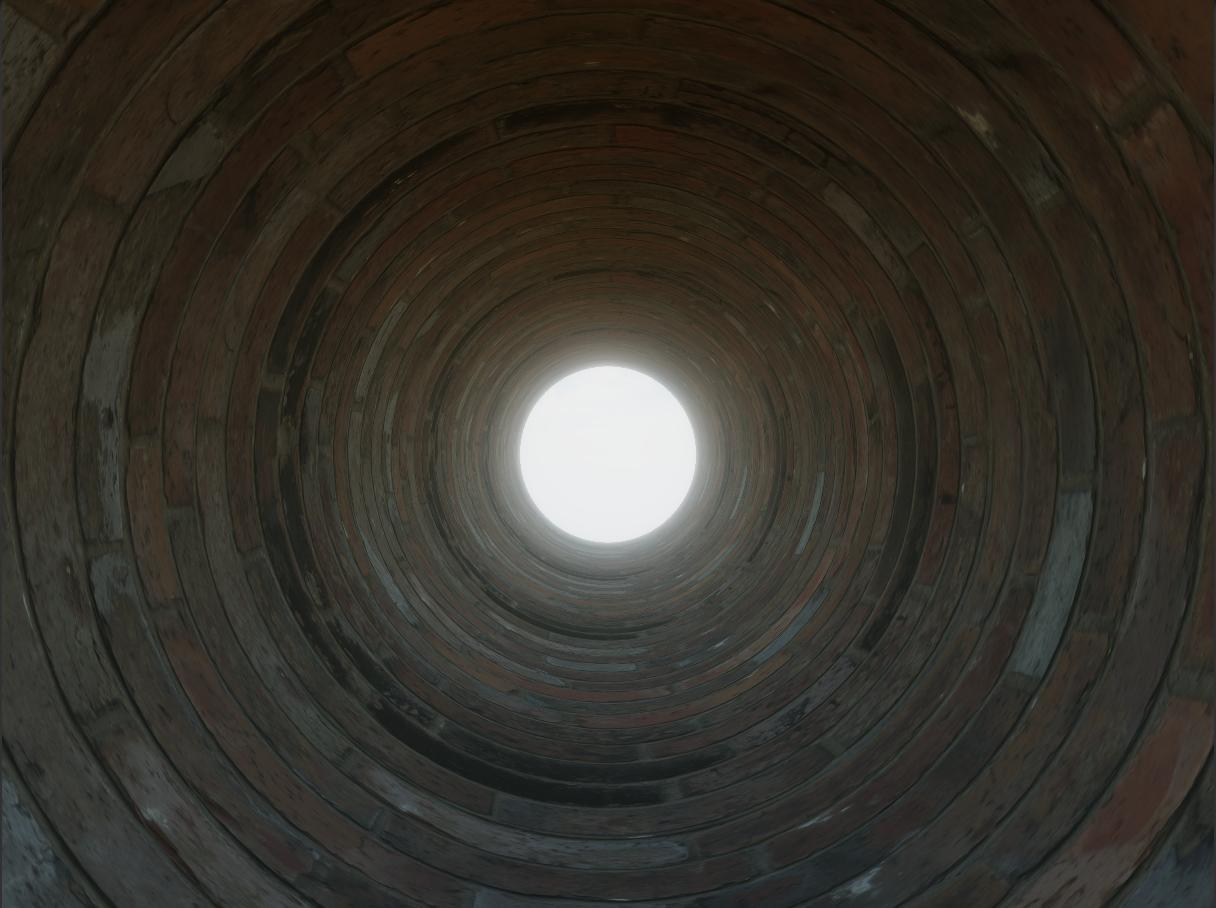}
		\caption{}
	\end{subfigure}  
	\begin{subfigure}[b]{0.22\textwidth}
		\includegraphics[width=\textwidth]{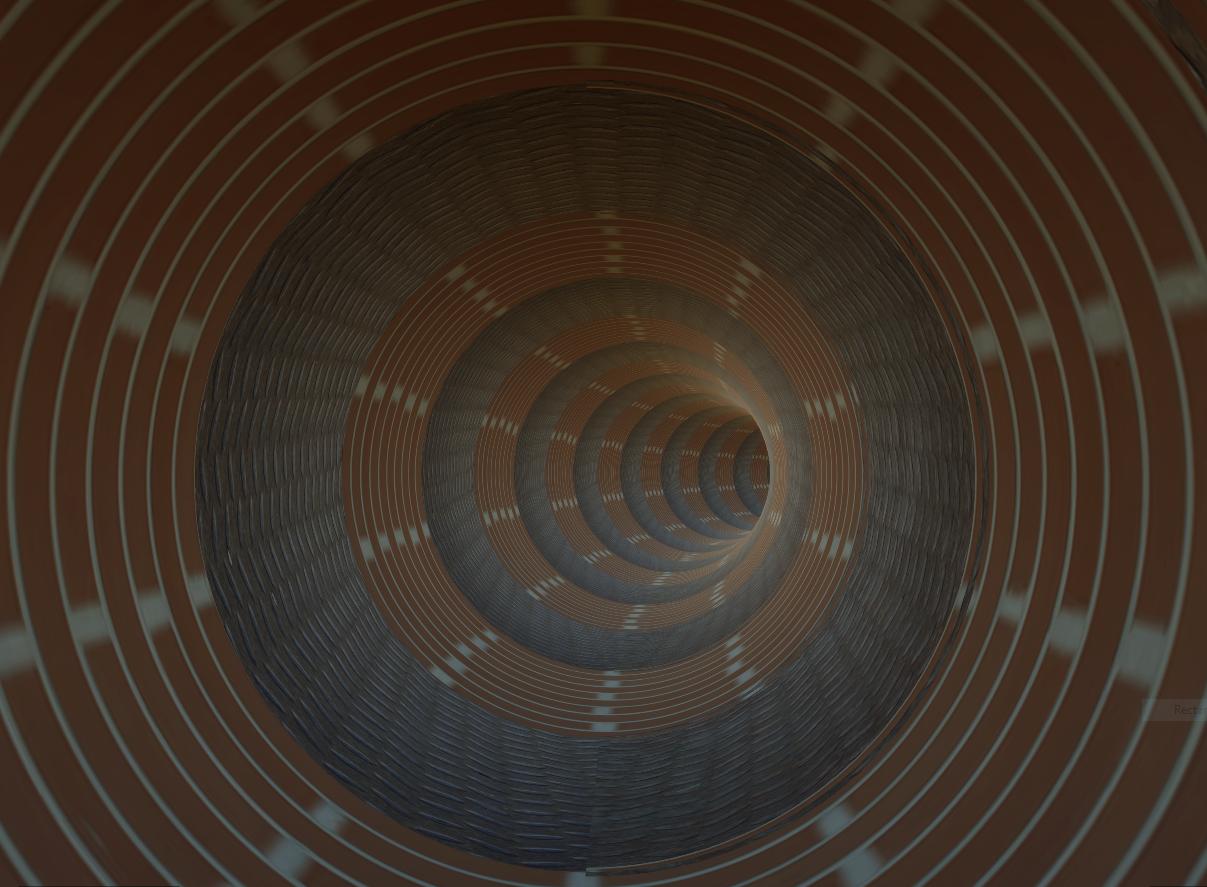}
		\caption{}
	\end{subfigure}
		\begin{subfigure}[b]{0.45\textwidth}
		\includegraphics[width=\textwidth]{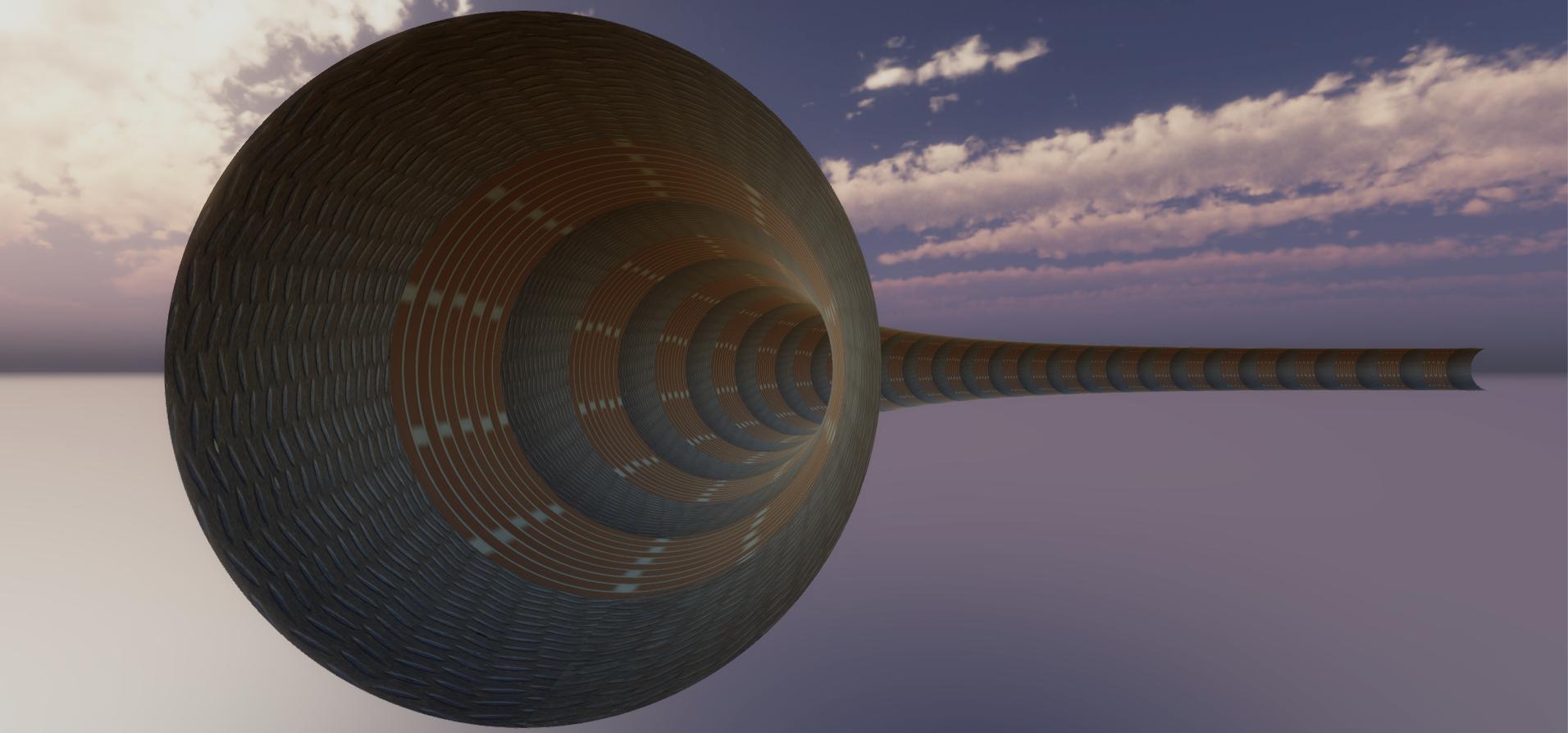}
		\caption{}
	\end{subfigure}  
		\begin{subfigure}[b]{0.45\textwidth}
		\includegraphics[width=\textwidth]{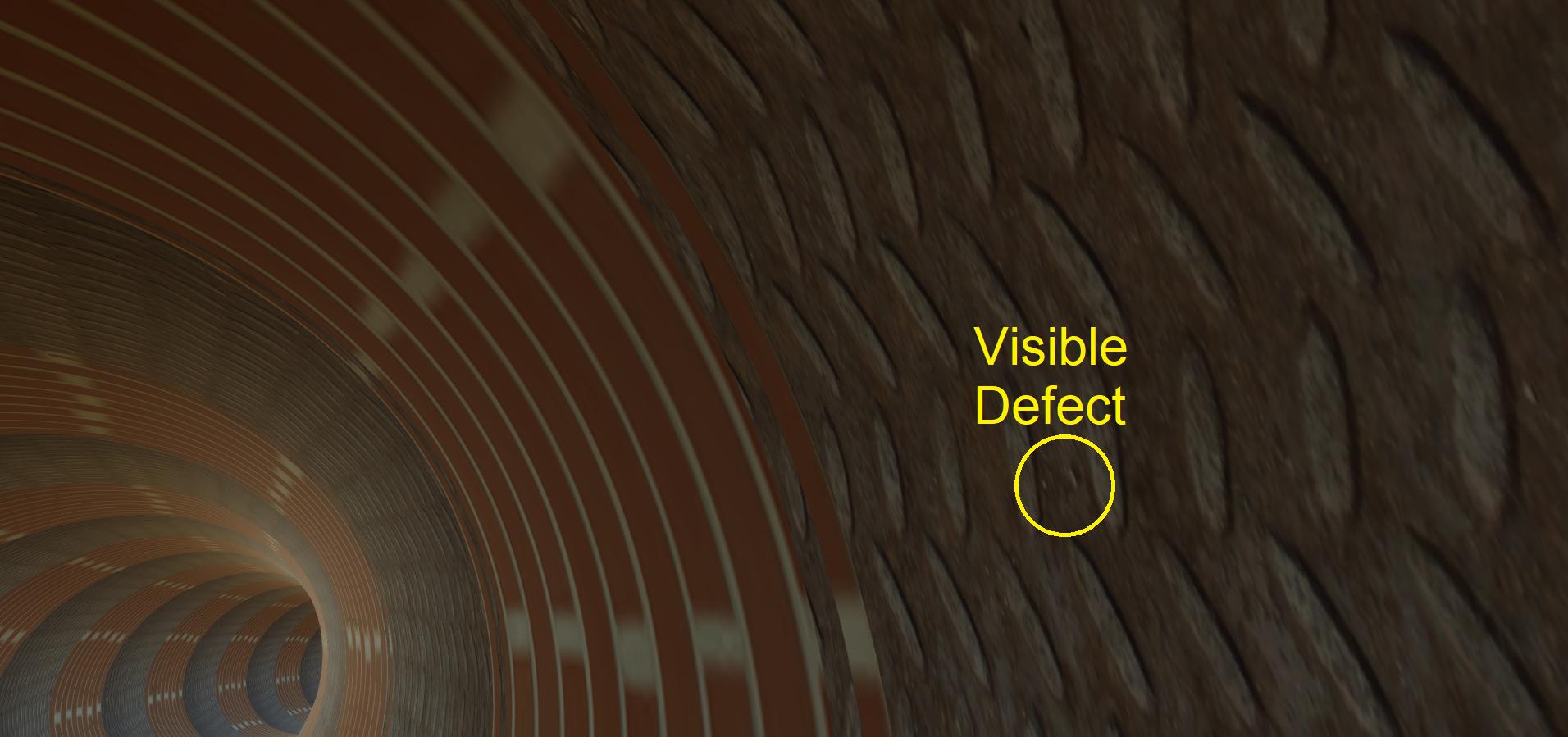}
		\caption{}
	\end{subfigure}  
  \caption{A hollow cylinder is created in Blender and UV mapped. This cylinder is imported in unity to create both (a) straight and (b,c) curved tunnels based on the user requirements. (d) The user can move freely inside the tunnel and move closer to the boundaries to carefully inspect for any damage in the tunnel.}
\label{fig::unity_display}
\end{figure}
In the baseline stitch, we assume that the camera does not suffer from any jitter or sideways movement and blindly trust the initially planned rotation and translation per frame. In our second approach, we use groundtruth measurements to perform image stitching. The results of the baseline and groundtruth stitch are shown in Fig.~\ref{fig::SfM_stitch_seattle}. The baseline stitch, understandably, fails to provide us with a good stitch of the simulated tunnel. This validates our framework and showcases that we can obtain good results for stitching and visualizing long underground tunnels as long as we are able to obtain a good estimation for camera pose per image. This cylindrical projection stitch can be wrapped around a cylinder in the Unity engine to provide an immersive $3$D display of the panoramic view for users.

\noindent \textbf{Visualization in Unity}: We display the cylindrical images of the scene rendered as described in previous sections in Unity. We envision our system as a ``fly-through" of the scene. The user controls the camera position using keyboard and mouse and just like a first-player shooter game, can float around in the $3$D scene freely. This enables the user to not only view the rendered tunnel images but also move closer to the areas where the user suspects faults in the tunnel. Fig.~\ref{fig::unity_display} shows our setup for straight and curved tunnels. The user provides a text file with the curve of tunnel and the cylinders for the curved tunnel are rendered when the user starts the application. Video demos for Unity simulation and user inspection can be seen  \href{https://www.youtube.com/watch?v=g17TKL57gjE}{here} and \href{https://www.youtube.com/watch?v=sGoxhKr4pjw}{here}.

\end{section}

\begin{section}{Conclusion}\label{sec::conclusion}
We presented a simple and accurate system to capture images of a given scene using a spirally moving camera system and display the panoramic stitched images in unity for an interactive $3$D display. The presented method excels in scenes where prior geometrical information is available. This allows us to project the images in $3$D and warp them onto a unit cylinder to obtain unit cylindrical images of the scene. This approach can be easily extended to other commonly found geometries such as underpasses, rooms, and train tracks.

In future, we plan to extract the geometrical information such as tunnel radius, curved angle of pipes automatically using depth sensors mounted on the UAV. We also intend to test our framework on real dataset in the future. 
\end{section}
\section*{Acknowledgment}

This research grant is supported by the Singapore National Research Foundation under its Environmental $\&$ Water Technologies Strategic Research Programme and administered by the PUB, Singapore's National Water Agency.

\bibliographystyle{IEEEtran}
\bibliography{thesisrefs}

\end{document}